# CHAINTRIX: A multi-pipeline LLM-augmented framework for automated smart-contract security auditing

Gabriela DOBRIȚA, Simona-Vasilica OPREA, Adela BÂRA
Bucharest University of Economic Studies, Department of Economic Informatics and Cybernetics, no.6 Piața Romană, Bucharest, 010374, Romania, *Corresponding author: simona.oprea@csie.ase.ro

**Abstract:** Smart-contract exploits have caused billions of USD in cumulative losses, yet audits remain expensive and slow. Automated tools have emerged to close this gap, but each class has a characteristic failure mode. Static analyzers report findings that frequently fail manual triage at high rates, while large language models (LLMs) hallucinate findings that contradict the source code. Thus, we propose Chaintrix, an end-to-end auditing framework whose central architectural commitment is that every LLM-generated claim must be discharged against a deterministic structural contract representation. We introduce a Cross-Contract Interaction Model (CCIM) that parses Solidity into a structured map of function-level reads, writes, modifiers and resolved cross-contract calls. CCIM serves as the substrate against which all 12 of Chaintrix's deterministic signal engines and the parallel LLM audit pipelines operate. A staged false-positive-reduction pipeline, terminating in a Structural Verdict Engine (SVE) that applies deterministic structural checks against parsed code, filters the merged finding set, with selected high-confidence findings further validated through symbolic execution and fuzz testing. We evaluate Chaintrix on EVMbench, the smart-contract security benchmark by OpenAI, Paradigm, OtterSec. Chaintrix detects 86 of 120 high-severity vulnerabilities (71.7% recall), with 25 audits scoring 100% recall, placing Chaintrix 26 percentage points above the strongest frontier-model baseline.

## 1. Introduction

Smart-contracts now custody hundreds of billions of dollars across decentralized finance, bridging and on-chain governance. The economic stakes have grown faster than the supply of qualified human auditors: professional audit firms charge between USD 50,000 and USD 500,000 per engagement, with lead times measured in weeks, while the rate of new contract deployment continues to outpace audit capacity by orders of magnitude. The consequences of this gap are quantifiable: over USD 3.8 billion was lost to smart contract exploits in 2022 alone [1], with several individual incidents exceeding USD 100 million in extracted value. The economic incentive to automate part of the auditing process is therefore both clear and urgent [2]. Existing automated approaches fall into two broad categories, each with characteristic limitations. Pattern-based static analyzers such as Slither [3], [4] achieve high recall on a fixed set of syntactic vulnerability patterns (reentrancy, uninitialized state, missing zero-address checks, weak randomness), and symbolic-execution tools such as Mythril [5] provide path-sensitive detection bounded by exploration depth and timeout constraints.

Both classes share a common limitation: empirical evaluations report individual-detector false-positive rates exceeding 90% on benchmark datasets [6] and overall pipeline rates of 50–70% on real protocol code [7], burying genuine findings in noise that human reviewers must manually triage. LLMs demonstrate strong code comprehension [8], [9] and can reason about cross-function semantics, but they hallucinate findings [10]. For instance, asserting that a function lacks access control when an onlyOwner modifier is plainly visible or claiming reentrancy in a contract that uses nonReentrant. The two failure modes are not symmetric and cannot be addressed by simply combining outputs: static-analyzer noise and LLM hallucination compound rather than cancel.

The fundamental challenge is that no single methodology offers both grounding and reasoning [11]. Static analysis is grounded in parsed code structure but cannot reason about protocol intent. LLMs reason about intent but lack a ground-truth anchor. Formal methods provide mathematical proof but require specifications that, in practice, no one writes. The architectural gap that follows is the central concern of this paper. It makes three contributions:

(1) *A Cross-Contract Interaction Model (CCIM)* that parses Solidity into a structured map of function-level reads, writes, modifiers and resolved cross-contract calls. CCIM serves simultaneously as evidence for LLM prompts and as ground truth for verifying LLM claims, providing the deterministic substrate that anchors all subsequent stages of the pipeline.

(2) *A multi-pipeline architecture that runs an interaction-driven (ID) LLM audit and a dossier-driven (DD) phased* audit in parallel over the shared CCIM substrate, merged through cross-pipeline root-cause deduplication and optionally extended with formal verification.
(3) *A multi-engine deterministic signal layer* combining in-house detectors: Inferred Typestate/Precondition Consistency (ITPC), Commit-Invalidate-Revoke (CIR), Cross-Contract Parameter Type Inference (CCPTI), Bugs-as-Deviant-Behaviour Pattern Mining (BPM) [12], Integration Risk Analyser (IRA), a novel Boundary Value Analyser (BVA) and a battery of pattern-specific custom detectors (signature, arithmetic, and assembly-level), with established external tools (Slither, Mythril). Engine outputs are fused into per-function risk dossiers that anchor LLM prompts in structural evidence and provide independent corroboration signals during triage.

The following two contributions derive from the previous ones:

(4) A five-stage false-positive-reduction pipeline composed of (i) deterministic claim verification against parsed structure, (ii) programmatic structural filters, (iii) evidence-packaged LLM re-verification, (iv) root-cause-based clustering with confidence scoring, (v) a SVE that applies eight deterministic structural checks against parsed code, followed by semantic LLM re-verification.
(5) An empirical evaluation on EVMbench [13] Detect-mode (120 high-severity vulnerabilities across 40 audits), reporting per-audit and aggregate recall and the distribution of perfect-score audits.

The novelty of Chaintrix lies in adopting representation-level verification rather than prompt-level coupling, using a deterministic cross-contract interaction model to anchor, deduplicate and verify audit findings, including those generated by LLMs

**2. Literature review**

Smart contracts have become a core building block of blockchain ecosystems, enabling decentralized and self-executing agreements without the need for intermediaries. Their adoption has expanded beyond cryptocurrency transactions into business networks, decentralized finance, IoT, logistics, governance and other digital infrastructures. However, the immutability of deployed contracts means that implementation errors may directly lead to security breaches and substantial economic losses. In this context, smart contract auditing has become essential for ensuring transparency, trust, and operational security in blockchain-enabled environments [14]. Comparative studies have shown that smart contracts differ significantly from traditional software systems in terms of privacy, communication models, update mechanisms and threat exposure. Various blockchain platforms, including Ethereum, Bitcoin, Stellar and others, exhibit unique attack surfaces, motivating the development of platform-specific taxonomies of vulnerabilities and security countermeasures [15]. To further understand the evolving threat landscape, comprehensive surveys have systematically classified smart contract vulnerabilities and reviewed existing security analysis techniques. These studies summarize both manual and automated approaches, including expert review, decompilers, semantic analysis frameworks, anomaly detection and scalable auditing solutions [16].

Large-scale empirical analyses of deployed Ethereum contracts further confirm that vulnerabilities remain widespread in practice. Static code inspections have revealed recurring issues such as arithmetic errors, unsafe external calls and logic flaws, emphasizing that many security weaknesses persist even in publicly deployed contracts [17]. Among known vulnerability classes, reentrancy remains one of the most damaging attack vectors in Ethereum ecosystems. Beyond traditional detection methods, recent work has examined reentrancy from an adversarial economics perspective, modeling attacker behavior under gas costs, execution constraints and risk considerations [18].

Systematic reviews of machine learning applications in smart contract security reveal a clear transition from rule-based detection toward more sophisticated architectures, including convolutional, recurrent, and graph-based neural networks. These studies emphasize that combining multiple data representations and learning paradigms can significantly enhance detection performance [19]. Building on this trend, transformer-based approaches have been introduced for smart contract security analysis. One representative framework combines BERT, bidirectional LSTM networks and attention mechanisms to extract semantic patterns from contract opcodes [20]. To better capture program structure, graph-based representations have also been proposed. One approach constructs semantic graphs from abstract syntax trees enriched with control-flow and data-flow relationships, then applies edge-aware residual graph convolutional networks for function-level vulnerability detection. The results demonstrate that structural code representations significantly improve the ability to identify vulnerable contract behaviors [21]. Other studies further integrate graph neural networks with expert-defined security knowledge. By combining control-flow graphs, temporal message propagation and handcrafted vulnerability patterns, these hybrid systems achieve strong detection performance across multiple vulnerability classes, including reentrancy, timestamp dependence and infinite loops [22].

Beyond static and structural analysis, dynamic testing approaches have also gained attention. Multi-Agent Systems combined with Deep Reinforcement Learning have been proposed for fuzzing-based security assessment. These frameworks address challenges such as state-space explosion and real-time exploit discovery, demonstrating the potential of adaptive multi-agent exploration for strengthening smart contract testing pipelines [23]. More recently, large language models have been investigated for automated smart contract auditing. Empirical studies evaluating ChatGPT show that LLMs can achieve high recall in vulnerability detection tasks, indicating strong semantic reasoning capabilities. However, they also suffer from inconsistent precision, hallucinated findings and context-length limitations, which reduce their reliability for production-grade auditing without additional verification mechanisms [24]. To address these shortcomings, hybrid intelligent frameworks have begun combining optimization algorithms with anomaly detection techniques. One example integrates genetic algorithms with isolation forests to improve detection efficiency while reducing false positives [25].

Furthermore, [26] introduced Securify, a security analyzer that encodes a Solidity contract's control- and data-flow dependence graph in stratified Datalog and then checks compliance and violation patterns expressed in a domain-specific language. The idea is to mirror each security property by two complementary patterns: a compliance pattern whose match implies satisfaction of the property and a violation pattern whose match implies its negation. Behaviors matching neither are reported as residual warnings. This three-way classification lets a Datalog solver discharge the proof obligation in seconds, and the publicly released tool has been used to analyse over 18,000 contracts. The authors motivate this design as a response to the dual failure mode they attribute to symbolic-execution baselines such as Oyente and Mythril: under-approximation that misses critical violations and imprecise domain-specific modelling that produces false positives. Securify's stated limitations are different in kind: no numerical reasoning, an assumed reachability of analysed instructions and warnings that still require manual triage.

Also, the first automated verifier, VerX, capable of proving custom functional properties of Ethereum contracts was presented. It combines three techniques: (i) reduction of temporal property verification to reachability checking, (ii) a symbolic execution engine tuned for a practical fragment of EVM code, and (iii) delayed predicate abstraction that interleaves symbolic execution within transactions with predicate abstraction at transaction boundaries. Additionally, [27] presented PropertyGPT, the most direct architectural comparison to Chaintrix among published LLM-augmented auditing systems. It addresses the specification bottleneck identified by VerX: it indexes a corpus of 623 human-written formal properties (invariants, pre/post-conditions, rules) by embedding their associated code into a vector database and uses retrieval-augmented generation with GPT-4 to synthesize customized properties for previously unseen contracts. To make the generated properties usable, the authors solve three sub-problems: compilability, appropriateness and verifiability, each addressed by a dedicated mechanism: compiler-driven iterative revision, weighted multi-dimensional similarity ranking and a final formal prover. The reported results are an 80% recall in property generation against a held-out ground truth, detection of 9 of 13 historical Common Vulnerabilities and Exposure (CVE) and 17 of 24 attack incidents (26/37 combined), twelve zero-day discoveries and $8,256 in disclosed bug-bounty rewards. PropertyGPT and Chaintrix occupy adjacent but distinct positions in the design space: PropertyGPT generates inputs for a downstream prover and depends on the existence of a curated property corpus elsewhere, whereas Chaintrix performs end-to-end discovery without any such corpus, anchoring every LLM prompt in the deterministic cross contract interaction model rather than in a code-similarity neighborhood. Also, GPTLens was introduced, an adversarial two-stage framework in which an LLM plays an auditor (high-temperature generation of candidate vulnerabilities) followed by a critic (discrimination on correctness, severity and profitability) [28]. Their motivating

discussion documents extreme precision shortfalls: single-digit percent on real-world DeFi contracts at moderate recall, for direct LLM auditing without structural anchoring. On their own evaluation of 13 CVE-tagged contracts, the two-stage critic lifts CVE-level hit ratio from 38.5% (one-stage) to 76.9%, showing that adversarial decomposition helps but leaves the precision floor unaddressed.

Hybrid systems combining static analysis with LLMs already exist in commercial and open-source form, but they couple the two stages loosely: the static tool's output flavors the LLM prompt, the LLM's output is reported directly and no shared structural model verifies either side's claims. Such systems compound the false-positive overhead of the static layer with the hallucination overhead of the LLM layer rather than reducing either. The research gap is architectural rather than algorithmic, each tool category producing signals of fundamentally different types, syntactic patterns versus semantic narratives, that require a unifying framework to correlate, verify and filter.

Unlike existing smart contract auditing systems that either statically flag syntactic patterns, formally verify user-provided properties or semantically reason over code using unconstrained LLMs, Chaintrix proposes a different approach for smart-contract automated auditing architecture in which all heterogeneous security signals, deterministic, semantic and formal, are anchored to a shared cross-contract structural model and every LLM-generated vulnerability claim must be deterministically discharged before being accepted.

## 3. Methodology

### 3.1 System overview

Design principles embodied in the architecture are presented in the Supplementary material. Chaintrix accepts a Solidity repository or a single contract and emits a structured report of triaged findings. The system is implemented as an asynchronous pipeline orchestrated by a single CLI entry point; on submission, a repository ingestion stage classifies files by role (source, test, script, interface, library), resolves Forge remappings and dependency paths, selects the in-scope contract set and produces a concatenated audit source with a file-offset map that preserves line-accurate provenance for every later finding. Once the source is normalized, the pipeline executes in three macro-phases. First, a deterministic phase parses the source into the CCIM and runs the deterministic signal engines synchronously over the parsed structure. The engine outputs are fused into a single per-function signal record. Second, two LLM audit pipelines (ID and DD) execute concurrently over the shared signal record, each producing an independent finding set. Third, the two finding sets are merged through cross-pipeline root-cause clustering and then passed through the staged false-positive-reduction taxonomy, which produces the final report.

Figure 1 depicts the high-level architecture. Stage 1 resolves scope and compiler version; Stage 2 builds the CCIM and runs the deterministic signal engines, fused into a per-function signal record. Stage 3 runs the two complementary LLM audit pipelines concurrently over that shared signal record. The DD pipeline (Phases A–E) audits each function in five sequential phases. The ID pipeline (Stages1–5, red team review and chain synthesis) verifies structural-engine hypotheses, scans cross-function interactions, surveys the protocol at a systemic level, micro-scans individual functions, re-verifies all candidate findings and synthesizes multi-step exploit chains from the surviving set. The two pipelines feed the cross-pipeline root-cause merge, which deduplicates findings and assigns cross-pipeline corroboration scores. Stage 4 applies the remaining four reduction stages: Structural Claim Verification, Root-Cause Clustering and Triage, the two layers of the SVE and emits the Final Triaged Report.

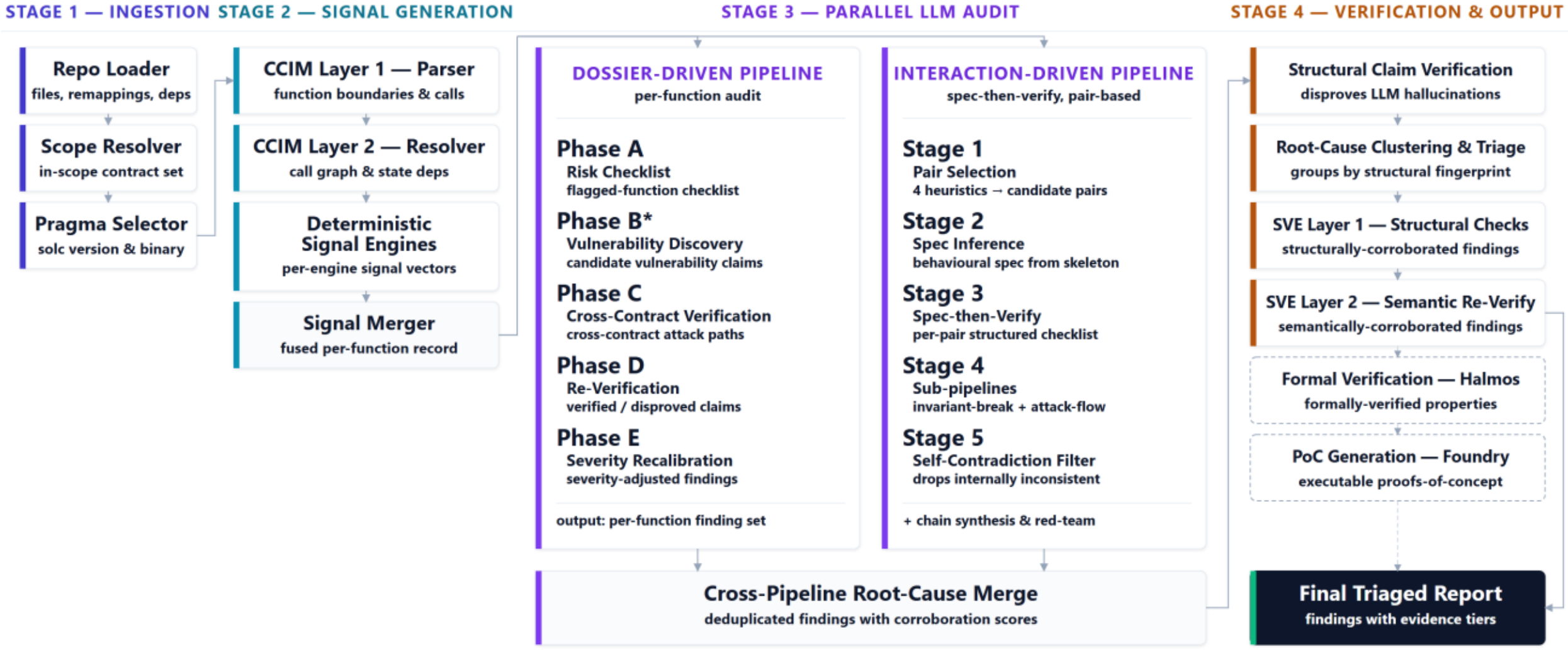

Figure 1. Chaintrix end-to-end architecture

## 3.2 System architecture

### 3.2.1 Cross-contract interaction model

The CCIM is the system's deterministic ground-truth layer: a regex-only parser that compiles the entire codebase, in under a second, into a single structured object that every downstream verifier checks the LLM against. Its atomic unit is a per-function record carrying the function's visibility, mutability, modifier chain, parsed access guards, the storage variables it reads and writes, the external calls it issues (each tagged with its target, method, and source line), flags for token approvals and fund movement, and the source span used for citation. Its defining mechanism is target resolution: a bare cross-contract call expression is not actionable until the system knows which concrete contract its target variable points to, so the resolver builds this mapping from state-variable type declarations and walks the inheritance graph from interface types to their implementing contracts. Every resolved edge in the resulting call graph therefore identifies a concrete caller-contract and callee-contract pair rather than a variable name. State writes, reads and fund movements are then propagated through intra-contract internal calls to a fixpoint, so a public entry point inherits everything its helpers transitively do.

Let $\mathcal{C}$, $\mathcal{F}$, $\mathcal{V}$ denote the universes of in-scope contracts, functions and storage variables. Every function belongs to exactly one contract, captured by the ownership map:

$$c : \mathcal{F} \to \mathcal{C}. \tag{1}$$

For each function $f \in \mathcal{F}$, the parser emits a structured record:

$$rec(f) \;=\; \langle\, vis(f),\; mut(f),\; M(f),\; G(f),\; R(f),\; W(f),\; X(f),\; \tau(f),\; src(f) \,\rangle, \tag{2}$$

where $vis(f) \in \{\text{public,external,internal,private}\}$ is visibility; $mut(f) \in \{\text{view,pure,payable,nonpayable}\}$ is mutability; $M(f)$ is the modifier chain; $G(f)$ is the set of parsed access guards (e.g. require(msg.sender == owner)); $R(f), W(f) \subseteq \mathcal{V}$ are direct storage reads and writes; $X(f)$ is the set of external-call sites; $\tau(f) \in \{0,1\}$ is the fund-transfer flag; and $src(f) = (\ell_{\text{start}}, \ell_{\text{end}})$ is the source span used for citation.

Each call site $x \in X(f)$ carries:

$$x \;=\; \langle\, target(x) \in V, method(x), line(x) \,\rangle, \tag{3}$$

where $target(x)$ is the storage variable holding the callee's address and $method(x)$ is the invoked function selector parsed from the source. The defining mechanism of Layer 1 is the resolution map:

$$\rho : \mathcal{V} \to \mathcal{C} \cup \{\bot\}, \tag{4}$$

built from the state-variable type map $type : \mathcal{V} \to \mathcal{T}$ and the inheritance partial order $\preccurlyeq_H$ on $\mathcal{C}$ (read $I \preccurlyeq_H c$ as "$c$ implements interface $I$"):

$$\rho(v) \;=\; \begin{cases} c & \text{if } \exists\, c \in \mathcal{C} : type(v) = I \;\wedge\; I \preccurlyeq_H c \;\wedge\; c \text{ is concrete}, \\ \bot & \text{otherwise.} \end{cases} \tag{5}$$

A call site $x$ is *actionable* if $\rho(target(x)) \neq \bot$. The cross-contract call graph $G = (\mathcal{F}, E)$ has an edge for every actionable call site whose resolution lands on a concrete callee function:

$$(f,g) \in E \iff \exists\, x \in X(f) : \rho(target(x)) = c(g) \ \wedge\ method(x) = g. \tag{6}$$

Each edge therefore identifies a concrete *caller-contract/callee-contract* pair $(c(f), c(g))$ rather than a symbolic variable name. Lifting $E$ from the function level to the contract level gives the contract-edge predicate

$$E_{\mathcal{C}}(c_1, c_2) \iff \exists\,(f,g) \in E : c(f) = c_1 \ \wedge\ c(g) = c_2, \tag{7}$$

used by the trust model. Let $I(f) \subseteq \mathcal{F}$ be the set of intra-contract internal calls of $f$. Direct reads, writes and transfers are propagated through $I$ to a least fixpoint:

$$\begin{aligned} R^\star(f) &= R(f) \ \cup \bigcup_{g \in I(f)} R^\star(g), \\ W^\star(f) &= W(f) \ \cup \bigcup_{g \in I(f)} W^\star(g), \\ \tau^\star(f) &= \tau(f) \ \vee \bigvee_{g \in I(f)} \tau^\star(g). \end{aligned} \tag{8}$$

Every public entry point thus inherits the storage and fund-movement footprint of every helper it transitively invokes. Layer 2 lifts this graph into a security-aware view. A state-dependency map records who writes and who reads each storage variable and how the reads are used; a variable that is admin-rotatable and simultaneously used as a call target or approval recipient is flagged as a rotation risk. A trust model aggregated per directed contract pair records what the caller assumes versus what the callee enforces, with mismatches surfaced as trust gaps and bidirectional patterns flagged as callbacks. For each function $f \in \mathcal{F}$, let:

$$approvals(f) = \{ v \in \mathcal{V} \mid v \text{ appears as a recipient argument of an approve or safeApprove call site in } X(f) \} \tag{9}$$

denote the set of storage variables passed as token-approval recipients in $f$. For each storage variable $v \in \mathcal{V}$, Layer 2 records its writers, readers and "consumers" (functions that use $v$ as a call target or approval recipient):

$$\begin{aligned} \delta_W(v) &= \{ f \in \mathcal{F} \mid v \in W^\star(f) \}, \\ \delta_R(v) &= \{ f \in \mathcal{F} \mid v \in R^\star(f) \}, \\ uses(v) &= \{ f \in \mathcal{F} \mid (\exists x \in X(f) : target(x) = v) \ \vee\ v \in approvals(f) \}. \end{aligned} \tag{10}$$

A function is *admin-callable* if its parsed guards match a known role-check pattern from a fixed catalogue $\mathcal{P}_{\text{role}}$ recognized by the parser (e.g. onlyOwner, onlyRole(...), require(msg.sender == owner), OpenZeppelin AccessControl modifiers):

$$admin(f) \iff G(f) \cap \mathcal{P}_{\text{role}} \neq \emptyset. \tag{11}$$

A storage variable is *rotation-risky* if it is admin-rotatable and simultaneously consumed:

$$rot(v) \iff (\exists\, f \in \delta_W(v) : admin(f)) \ \wedge\ (uses(v) \neq \emptyset). \tag{12}$$

For each directed contract pair $(c_1, c_2)$ with $E_{\mathcal{C}}(c_1, c_2)$, define the *assumed* and *enforced* predicate sets:

$$\begin{aligned} assumes(c_1, c_2) &= \bigcup_{(f,g) \in E,\ c(f) = c_1,\ c(g) = c_2} post(g), \\ enforces(c_2, c_1) &= \bigcup_{g \in \mathcal{F},\ c(g) = c_2} guard_{c_1}(g), \end{aligned} \tag{13}$$

where $post(g)$ is the set of post-conditions parsed from return and emit statements of $g$; and $guard_{c_1}(g)$ is the subset of $G(g)$ that gates calls originating from $c_1$. A *trust gap* is a containment failure:

$$trustgap(c_1, c_2) \iff assumes(c_1, c_2) \not\subseteq enforces(c_2, c_1). \tag{14}$$

A *callback* is a bidirectional edge at the contract level:

$$callback(c_1, c_2) \iff E_{\mathcal{C}}(c_1, c_2) \ \wedge\ E_{\mathcal{C}}(c_2, c_1). \tag{15}$$

The structural object emitted to all six downstream consumers (Figure2) is the tuple:

$$\text{CCIM} = \big\langle \{\text{rec}(f)\}_{f \in \mathcal{F}},\ \rho,\ \mathcal{G},\ \{R^\star, W^\star, \tau^\star\},\ \{\delta_W, \delta_R, uses, rot\},\ \{assumes, enforces, trustgap, callback\} \big\rangle. \tag{16}$$

Every downstream stage of Chaintrix reads from this tuple as its sole structural ground truth. The Structural Claim Verifier checks LLM claims against $\rho$ and $G(f)$. The SVE consumes $vis, mut, M(f)$, the DD and ID pipelines build per-function dossiers and interference-pair sets from $R^\star, W^\star, \delta_W, \delta_R$. The Halmos stage targets $\delta_W(v) \cup uses(v)$ for every $v$ with $rot(v)$ and the report generator emits $src(f)$ as line-range citations.

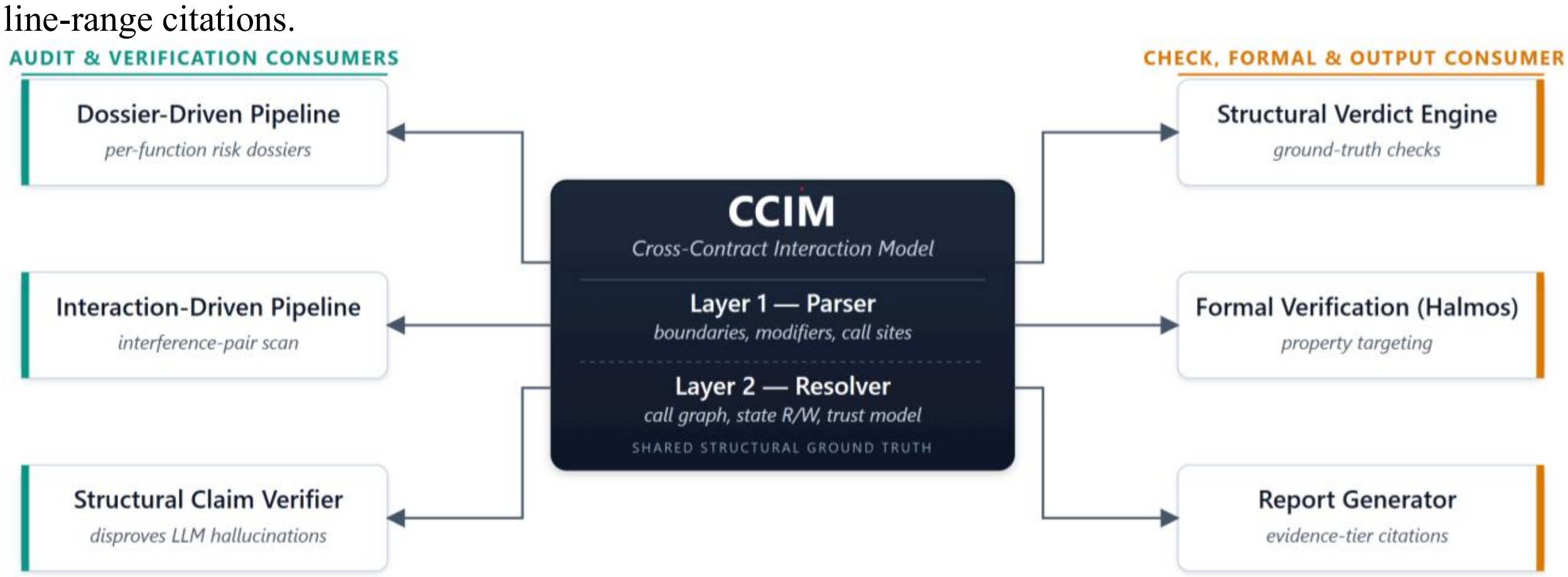

Figure 2. CCIM as shared ground truth

### 3.2.2 Deterministic signal engines

The deterministic engines run sequentially before any LLM call, producing the structural signal base that anchors every downstream prompt, as in Table 1. Each engine is invoked under per-engine exception isolation: if one fails, an empty fallback result is logged and the pipeline continues, so no single component can block the analysis. Engines relying on Slither share a single parsed repository representation, and all engine outputs are normalized into structured per-function risk signals containing source, severity and evidence metadata.

Table 1. Detection scope for deterministic engines

| Engine | Abbreviation | Detection target |
|---|---|---|
| **Inferred Typestate/Precondition Consistency** | ITPC | Functions whose inferred preconditions are inconsistent across call chains |
| **Commit-Invalidate-Revoke** | CIR | External commitments (approvals, delegations) that become stale after state changes |
| **Cross-Contract Parameter Type Inference** | CCPTI | Semantic type mismatches (block number vs timestamp) propagated through calls |
| **Bugs-as-Deviant-Behavior Pattern Mining** | BPM | Functions that deviate from majority patterns (guards, co-modifications, action-reaction) |
| **Integration Risk Analyzer** | IRA | Structured risk questions at external call boundaries for LLM follow-up |
| **Custom Detectors (x11)** | CUSTOM | Patterns missed by Slither (oracle staleness, div-before-mul, unsafe casts) |
| **Signature Detector** | SIG | Replay, forgery, and EIP-712 compliance issues |
| **Math Detector** | MATH | Precision loss, unsafe casts, unchecked-assembly arithmetic |
| **Assembly Detector** | ASM | Inline-assembly pitfalls (delegatecall forwarding, returndata handling) |
| **Slither** | SLI | Full Slither detector suite, normalized for signal-merger consumption |
| **Mythril** | MYT | Symbolic-execution counterexamples |
| **Boundary Value Analyzer** | BVA | Boundary, value-flow, rationality, formula and invariant analysis (this work) |

The BVA is the only fully novel engine in the suite; the remaining eleven are either re-implementations of established techniques (BPM follows Engler et al.'s "bugs as deviant behavior" methodology [12]; ITPC adapts inferred-typestate checking [29] to Solidity) or normalized integrations of existing tools (Slither, Mythril). BVA itself is composed internally of eight sub-analyzers, each run per in-scope (non-framework) contract: (i) value-flow extraction; (ii) boundary finding from explicit require and if numeric guards; (iii) rationality checking, which tests whether each bound is reachable and internally consistent with the value flows that cross it; (iv) locked-ETH detection (contracts that can receive Ether but expose no withdrawal path); (v) read-before-write analysis; (vi) formula-mismatch detection across paired functions (e.g. deposit and withdraw using inconsistent price formulas); (vii) small-scale symbolic evaluation of arithmetic; and (viii) invariant-consistency checking across functions that should agree on a shared property.

The signal merger sorts each engine's output by a five-level severity ranking (CRITICAL through INFO, higher denoting more urgent triage) and applies a global cap of 50 signals to bound the context window of the downstream LLM prompts. The merged output is a per-engine dictionary plus a statistics block; a formatter converts this into a single Markdown string with one section per signal source, injected directly into the LLM system prompts. In practice the per-engine volumes are heavily skewed: ITPC and IRA dominate the merged pool, while detectors with stricter triggering conditions (CIR, MATH, ASM) typically produce single-digit signal counts, so the 50-signal cap binds primarily on ITPC and IRA outputs.

### 3.2.3 DD pipeline

This pipeline bridges the deterministic signal layer and the LLM by reorganizing every engine's output around the unit the LLM will reason about: the individual function. For each non-interface function known to CCIM Layer 1, the dossier compiler builds a per-function record that begins with the function's structural facts copied from CCIM (visibility, mutability, modifiers, access guards, the storage variables it reads and writes, its external calls, its fund-movement flag, its source span and its raw body.

Risk items contributed by each upstream engine are then attached to the appropriate function: BPM violations to the offending function; IRA questions to the caller of each external call; CCIM Layer 2's rotation and trust-gap findings to every function that reads the affected variable or calls the affected callee; and custom-detector, Slither and Mythril hits to the function each finding cites (matched by name, with a line-range fallback for Mythril findings that report only a line number). Each risk item carries a source tag, an identifier, a one-line description, a confidence score and a line hint; within each dossier the items are sorted by confidence so the LLM sees the strongest signal first. A function is considered flagged when its dossier contains at least one risk item after all attachments, and only flagged functions are sent through Phase A. The DD phased audit pipeline executes in five phases (A-E) as in Figure 3.

*Phase A (risk checklist).* Phase A turns each flagged dossier into a per-function checklist and asks the LLM, item by item, whether each pre-identified risk is real. The system prompt instructs the model to act as a Solidity security auditor and to return for every checklist item a verdict (REAL, FALSE_POSITIVE, or UNCLEAR) backed by an evidence line citation. Four built-in false-positive rules are encoded directly in the prompt so the model never spends reasoning on them: arithmetic overflow on Solidity $\geq 0.8$ outside unchecked blocks, reentrancy on functions guarded by nonReentrant, missing access control when an admin modifier is in fact present and any claim of an "EVM race condition" (the EVM's atomic transaction model makes them structurally impossible). All flagged dossiers are verified concurrently. A confirmed item becomes a structured finding carrying a title, description, attack scenario, severity and the list of affected functions.

*Pass 1-Discovery Phases (B, B2, B3, B4, C).* Pass 1 runs five discovery phases in parallel, each viewing the codebase through a different analytical lens.

*Phase B (per-contract bottom-up).* Phase B performs contract-level semantic analysis using the structural evidence accumulated during earlier stages. Contracts are prioritized according to aggregated deterministic risk scores and analyzed for logic flaws, economic inconsistencies, state-corruption paths and protocol-level attack scenarios extending beyond the initially detected risks. The analysis incorporates cross-call context, previously verified findings and structural interaction metadata derived from CCIM. Additional phases perform invariant extraction, systemic attack-path analysis and adversarial red-team verification (included in Supplementary material).

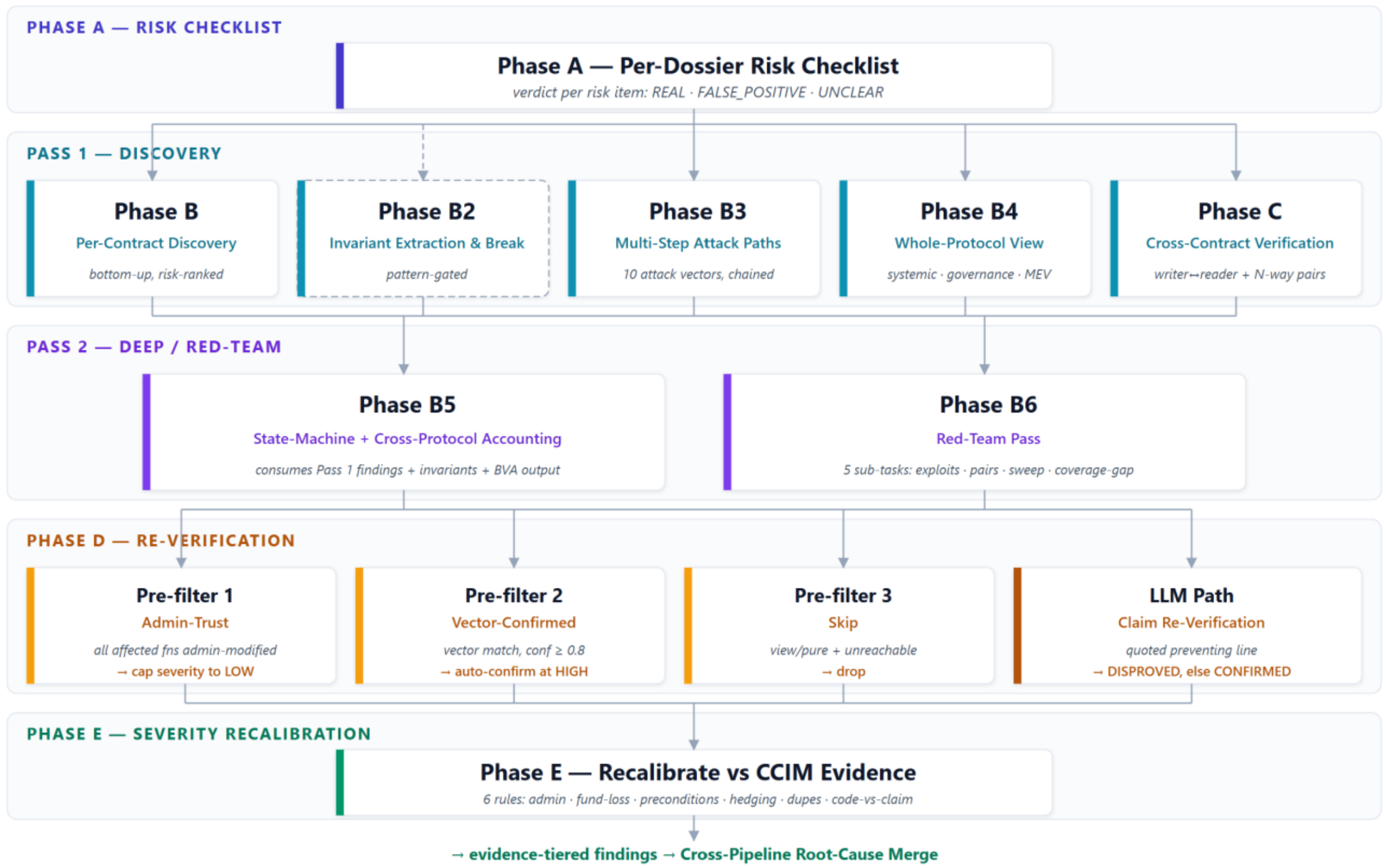

Figure 3. The DD phased audit

*Phase C (cross-contract verification)*. Phase C verifies the cross-contract interactions identified in CCIM Layer 2. From the state-dependency graph and the call graph, it constructs both pairs (a writer of a variable paired with each reader of the same variable or a caller paired with each callee) and N-way groups (three or more functions that all touch the same storage variable), so the model sees the entire interference set at once rather than one edge at a time. Each interaction returns a verdict (VULNERABLE, SAFE or UNCLEAR).

*Pass 2-Deep Phases (B5, B6)*. After Pass 1 findings are merged and deduplicated, two additional phases run with knowledge of what was already found. Additional phases regarding state-machine view and read team are presented in Supplementary material.

*Phase D (re-verification)*. Every finding from both passes goes through it. Three deterministic pre-filters route findings around the LLM whenever the answer can be decided structurally. Admin-trust captures findings whose every affected external function carries an admin modifier. They describe a trusted-admin action rather than a user-exploitable bug, so their severity is capped to LOW without any LLM call. Vector-confirmed captures findings that match a known attack-vector pattern with confidence at least 0.8, a non-empty evidence-line set and a proof trace of at least thirty characters. These are auto-confirmed at high confidence with no LLM cost. Graph-skip captures findings whose affected functions are all view or pure and absent from the call graph entirely. They cannot modify state, so the claim is unreachable by construction. The remainder reach the LLM through a claim-first verification protocol. The model is forced to (1) extract the finding's core claim in one sentence, (2) identify what specific piece of code would prevent that claim, (3) search the provided source for that specific prevention, and (4) issue a verdict: DISPROVED only if it can quote a concrete preventing line. The source block sent to the model is constructed by collecting every function the finding mentions and then expanding to include their callers and callees.

*Phase E (severity recalibration)*. It recalibrates severity using the actual access-control evidence extracted from CCIM rather than the finding's own claims. For each finding, the prompt receives the parsed visibility, modifier list, require-guard and state-write set of every affected function alongside the role hierarchy. The model applies six calibration rules and outputs both the recalibrated severity and a one-sentence justification that explicitly references the access-control evidence used.

**3.2.4 ID pipeline**

The ID pipeline is the second of the two LLM audit pipelines that run in parallel over the shared CCIM substrate and the merged deterministic-signal layer. Where the DD pipeline reasons one function at a time, the ID pipeline reasons one *interaction* at a time. Every audit unit consists of two related functions, an inferred behavioral specification of how they should compose and a structured checklist asking the LLM whether the code enforces or violates that specification. This decomposition is the architectural complement to the DD view as dossiers are strong on single-function logic errors but weak on emergent bugs that arise from how two functions interact through shared state. Pair-driven audit is strong on the latter and weak on the former. Running both pipelines concurrently and merging their findings by root-cause cluster is what gives Chaintrix its empirical recall on EVMbench.

The pipeline executes in five stages, followed by a red-team pass and an optional chain-synthesis stage.

*Stage 1-Pair candidate selection.* The pipeline begins not with the LLM but with four deterministic heuristics that nominate pairs of functions worth auditing together: (i) *Attention-based hotspot pairs* are read from the deterministic signal layer's attention scores, which rank function pairs by their estimated propensity to interact through shared storage or external calls; (ii) *Counter-pairs* are extracted by naming heuristics that match common protocol idioms (deposit↔withdraw, mint↔burn, lock↔unlock, stake↔unstake); (iii) *Shared-state pairs* are computed from CCIM Layer 2 by enumerating, for every storage variable $v$, the pairs $(f, g)$ with $v \in W^{\star}(f) \cap W^{\star}(g)$, functions that both write the same variable; (iv) *Deterministic-triage pairs* are nominated by the static engines themselves whenever two findings cite functions that share a parameter, a state read or a trust boundary. For low-risk contracts, a fifth, optional Claude-driven triage call adds suspicious interactions the four deterministic heuristics did not surface. The union of all five sources is deduplicated and ordered by source confidence so that the strongest signals are audited first when the downstream LLM context window is bound.

*Stage 2-Behavioural specification inference.* Each candidate pair is then sent through a behavioral-spec inference step whose central design decision is to look at the contract's *skeleton*, function signatures, NatSpec comments and module documentation, but not at the implementation. The model is asked to articulate, for the given pair, the expected lifecycle (in what order the two functions are intended to be called and under what conditions), the state variables they should agree on and the behavioral assumptions a correct implementation would have to preserve (for example: "withdraw must subtract the same amount that deposit added"). Because the spec is derived without seeing the implementation, the model cannot confabulate post-hoc justifications for whatever the code happens to do; the specification is forced to act as an independent oracle against which the implementation is then measured. This is the first pipeline's structural answer to the hallucination failure mode.

*Stage 3-Per-pair spec-then-verify.* It is the principal LLM-bearing stage. For each pair, the prompt carries the inferred spec from Stage 2, the source of both functions with their visibility, modifier and reentrancy-guard metadata copied from CCIM, the union of their preconditions inferred by ITPC and a structured seven-point checklist that the model must answer item-by-item: assumption enforcement (does the code ENFORCE or VIOLATE each specified assumption, with line citations), shared-state usage, value-flow trace, accounting consistency, invariant preservation across the pair, arithmetic safety (precision, overflow, unchecked blocks) and a final per-pair verdict. Each item must produce either an evidence citation (line number plus quoted code) or a concrete exploit trace. The prompt explicitly forbids invented attack narratives without a specific assumption violation. In parallel, every standalone function whose ITPC profile flags it as high-risk (for example, those handling value with unchecked arithmetic) is given its own focused audit slot outside the pair structure, so single-function bugs are not starved of attention by the pair-driven decomposition. Optional specialized sub-pipelines (*Stage 4*) extend analysis to invariants, authentication flows and cross-module interactions (Supplementary material).

*Stage 5-Self-contradiction filter and severity recalibration.* Before findings exit the pipeline, two cleanup steps run deterministically. A *self-contradiction filter* scans every finding's own evidence text for self-disproving language ("by design", "intended behavior", "not a vulnerability") and downgrades or removes findings whose own narrative refutes their verdict. This filter removes a measurable fraction of LLM noise without any further model calls. A *severity-recalibration* pass then re-grades the surviving findings against the parsed access-control evidence from CCIM rather than the finding's own claims, applying six explicit rules (admin-only paths capped to LOW unless they move user funds; findings with no concrete fund-loss path capped to MEDIUM; findings requiring three or more independent unlikely preconditions downgraded one level; hedged language without concrete attack steps capped to MEDIUM; root-cause duplicates marked and only the most impactful kept; CCIM evidence trusted over the finding's own access-control claim where they conflict). The output of Stage 5 is the ID finding set $F_i$ used by the cross-pipeline merge.

The ID pipeline's role is structural rather than additive, covering exactly the part of the vulnerability search space that a function-by-function decomposition cannot reach. Pair-based audit naturally surfaces stale-approval bugs, write-then-read races on shared state, accounting drift between paired update functions, lifecycle violations on multi-step state machines and trust-boundary mismatches across cross-contract calls, classes where the bug is not in any single function but in the *relationship* between two of them. Spec-then-verify provides a hallucination barrier for this class because the inferred spec is generated without sight of the implementation. Structural sub-pipelines extend the recall of the pair audit into invariant-shaped and auth-flow-shaped bugs that pair enumeration alone would miss, and the module dispatcher keeps wall-clock and token cost bounded on large repositories.

#### 3.2.5 Parallel execution and cross-pipeline merge

CCIM Layer 1 and the merged deterministic-signal layer form a shared substrate that is computed exactly once, on top of which the two audit pipelines launch concurrently. The synchronous deterministic engines are offloaded to a worker thread so that the asynchronous event loop is never blocked while the LLM phases of either pipeline are in flight, and the file-offset map is deep-copied per pipeline to prevent races on the per-pipeline annotations each one writes during execution.

Let $F_d$ and $F_i$ denote the finding sets produced by the DD and ID pipelines, respectively. After the per-pipeline ID renumbering step, the two sets are disjoint:

$$F_d \cap F_i = \emptyset. \tag{17}$$

We may therefore define the source-pipeline tag as a total function:

$$\pi: F_d \cup F_i \to \{D, I\}, \qquad \pi(f) = \begin{cases} D & \text{if } f \in F_d, \\ I & \text{if } f \in F_i. \end{cases} \tag{18}$$

We write $F = F_d \cup F_i$ for the merged finding set.

The unified set then enters structural claim verification and the root-cause triage stage as a single batch. Cross-pipeline deduplication is performed by the triage stage using the root-cause clusters it computes: whenever a cluster contains findings contributed by both pipelines, every member of that cluster is tagged *cross-pipeline* and is annotated with the matching identifiers from the other pipeline. This tag is consequential, contributing a +0.30 confidence bonus in the smart-triage scoring rule, reflecting the strong evidential value of two architecturally independent pipelines arriving at the same root cause from different decompositions of the search space.

The triage judge extracts a structural card (*sc*) from each finding and groups findings whose root cause, vulnerable function and impact agree. This induces a partition $\mathcal{D}$ of $F$:

$$\mathcal{D} \subseteq 2^F, \ \bigsqcup_{d \in \mathcal{D}} d = F, \ \forall d_1 \neq d_2 \in \mathcal{D}: d_1 \cap d_2 = \emptyset. \tag{19}$$

We write $sc: F \to \mathcal{D}$ for the function sending each finding to its (unique) cluster, so that:

$$f \in sc(f), \forall f \in F. \tag{20}$$

For each cluster $d \in \mathcal{D}$ the cross-pipeline indicator records whether *both* pipelines contributed at least one member:

$$\chi(d) \ = \ \mathbb{1}[\{\pi(f): f \in d\} = \{D, I\}]. \tag{21}$$

Note that $\chi(d) = 1$ requires $|d| \geq 2$, since the image of $\pi$ on a singleton is itself a singleton. Conversely, if $|d| = 1$ then $\chi(d) = 0$ automatically. Let $\text{conf}: F \to [0,1]$ denote the pre-merge confidence score assigned by smart triage. The cross-pipeline tag contributes an additive bonus $\beta = 0.30$ to every finding sitting inside a cross-pipeline cluster. Writing the post-merge score as $\text{conf}'$:

$$\text{conf}'(f) \;=\; \min\big(0.95,\ \text{conf}(f) + \beta \cdot \chi(sc(f))\big), \qquad \beta = 0.30. \tag{22}$$

The clipping reflects the reporting interval $[0.05,\ 0.95]$ used by the smart-triage scorer; without the clip a high base confidence plus the cross-pipeline boost could exceed 1. The choice $\beta = 0.30$ is the largest single positive term in the smart-triage score function. The cross-pipeline merge produces a tagged, partitioned and re-scored finding set:

$$\text{Merge}(F_d, F_i) \;=\; \langle\, F,\ \pi,\ \mathcal{D},\ \text{conf}' \,\rangle, \tag{23}$$

where $F = F_d \sqcup F_i$ is the disjoint union of pipeline outputs, $\pi\colon F \to \{D, I\}$ records the source pipeline of each finding, $\mathcal{D}$ is the $sc$-cluster partition of $F$, and $\text{conf}'\colon F \to [0.05,\ 0.95]$ is the post-merge confidence score augmented by the cross-pipeline indicator. Because both pipelines execute concurrently on the shared substrate, the end-to-end wall-clock time equals the cost of the substrate plus the slower of the two pipelines, rather than their sum. High-confidence findings may optionally undergo symbolic verification or executable proof-of-concept validation (as in Supplementary material).

### 3.3 False-positive reduction taxonomy

The defining failure mode of LLM-assisted auditing is *plausible but ungrounded* finding generation. A single audit prompt routinely produces tens of findings whose phrasing is convincing, but whose technical claims contradict the source. Chaintrix addresses this through what we term a *staged* verification taxonomy: five sequential filtering stages applied in cost order, where the central design contribution is not any individual stage but their *composition*: each stage removes a class of false positive that no other stage can decide. Deterministic refutation cannot reason about intent, LLM re-verification cannot cheaply check whether a function exists, structural filters cannot detect when an LLM's own evidence text contradicts its conclusion, clustering cannot disprove individual claims. Composed in cost order, with the cheapest decisive method first, the taxonomy is both economical and recall-preserving.

In *Stage 1,* every LLM finding is treated as a *testable proposition* about CCIM ground truth. The finding is classified into a small set of refutable claim types, missing access control, reentrancy, integer overflow on Solidity ≥ 0.8, "EVM race conditions" and checked against the parsed code: a claimed access-control vulnerability on a function bearing a verified access modifier is DISPROVED, with the modifier line cited as evidence. No LLM is consulted. This stage discharges the most structurally decidable hallucinations in milliseconds.

Three filter classes, in *Stage 2,* remove findings whose *structural locus* is unsound: generic centralization complaints lacking a concrete exploit path, findings whose own evidence text contains self-disproving language (*"by design"*, *"intended behavior"*), and findings citing functions that cannot be resolved against CCIM, the specific hallucination of plausible-sounding but non-existent function names.

*Stage 3* is the only LLM-bearing stage, running under tight evidence constraints. Three deterministic short-circuits: admin-trust, vector-confirmed and graph-skip, bypass the LLM whenever the verdict is structurally decidable. The remainder receive a function-bounded source extract, expanded along the call graph to include callers and callees, capped at a fixed character budget. The LLM is required to extract the finding's core claim, identify what code would prevent it, *quote* the preventing line if any and then issue a verdict: DISPROVED only when a concrete preventing line is quoted, never on assertion alone. This protocol is the principled inverse of free-form LLM auditing, the model being asked to find evidence *against* a finding, not to generate one.

Surviving findings in *Stage 4* are clustered by structural root cause: vulnerable function, abused state variable, attacker role and impact class, and assigned a confidence score combining severity, deterministic-engine corroboration and cluster characteristics. The decisive corroboration signal is cross-pipeline agreement (+0.30), triggered only when the same root cause is independently surfaced by both the DD and ID pipeline. Two architecturally independent decompositions converging on the same root cause is empirically rare for hallucinations and common for genuine vulnerabilities and is therefore treated as the strongest precision signal in the system.

A two-layer acceptance gate is designed in *Stage 5*. *Layer 1* applies eight deterministic ground-truth checks against the parsed code, each targeting a specific class of LLM hallucination, such as: claimed reentrancy on guarded functions, claimed fund-theft on functions that move no funds or claimed state corruption on view/pure functions. *Layer 2* re-verifies the survivors with a final evidence-packaged LLM call returning VERIFIED, DISPROVED or UNCERTAIN with supporting arguments. The two-layer split embodies the principle that recurs throughout the taxonomy: structurally decidable claims are decided structurally and only residual claims that genuinely require reasoning about intent are routed to the model. Figure 4 shows the resulting reduction funnel.

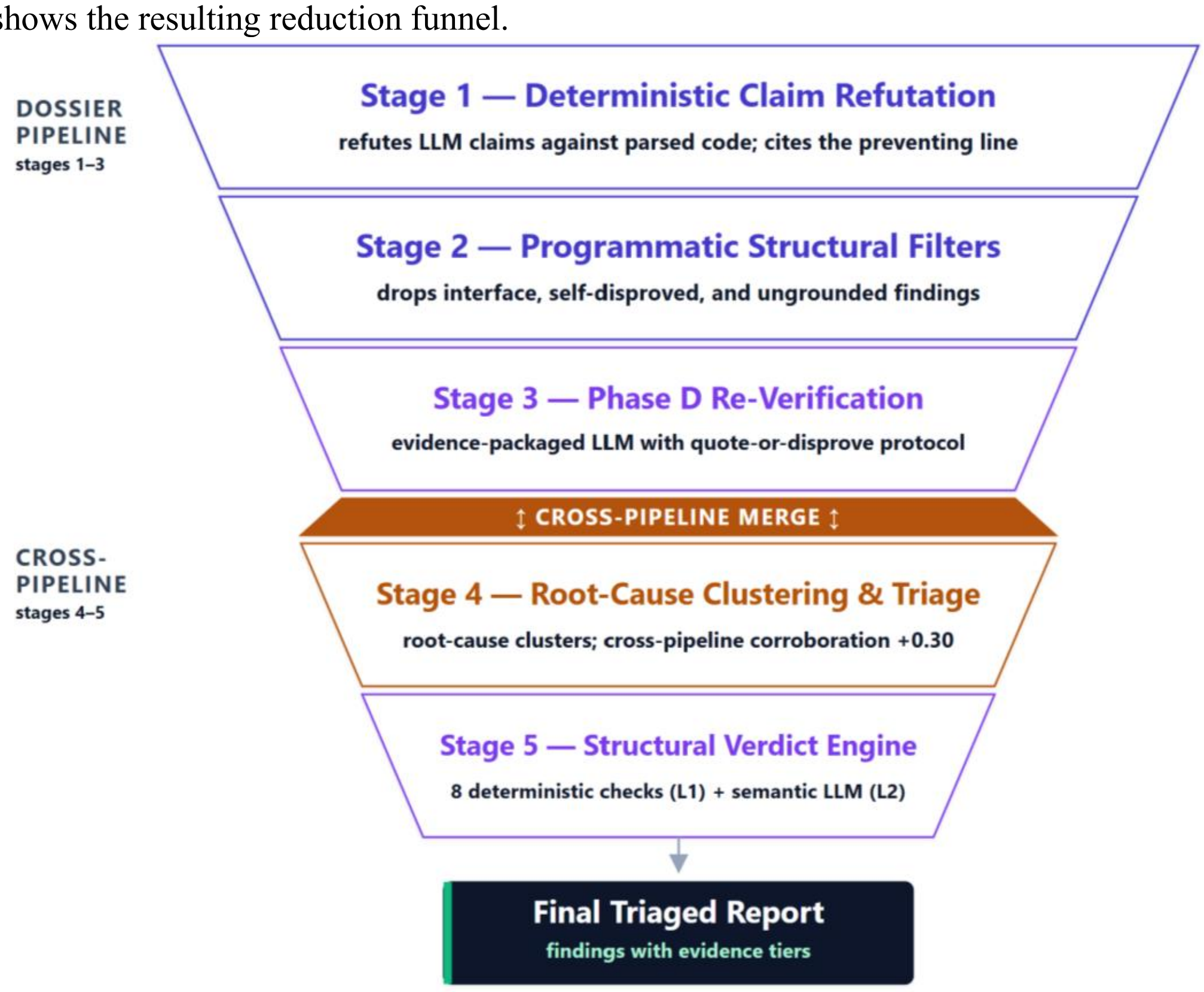

Figure 4. False-positive reduction funnel

### 3.4 Protocol coverage analysis

The previous section addressed precision, removing false positives from the finding set. This section addresses the dual concern of recall: *did the system actually examine the relevant attack surface?* A vulnerability that goes unfound is invisible to the false-positive funnel, so precision metrics alone are blind to it. Chaintrix mitigates this systemic recall gap with two complementary coverage mechanisms: one that reasons over *bug classes* (*what kinds of vulnerabilities should we be looking for?*) and one that reasons over *functions* (*which parts of the code did we actually look at?*). The two mechanisms are deliberately orthogonal. The first catches gaps along the bug-class axis even when every function has been examined, the second catches gaps along the function axis even when every bug class is in scope.

#### 3.4.1 Bug-class coverage and gap reporting

A structured knowledge base defines twenty protocol-feature categories: lending, staking, governance, AMM, oracle integration, auction, bridge, vault accounting and others, each populated with named bug classes and detection heuristics curated from the audit literature. The coverage analyzer first identifies which features are present in the target protocol from CCIM evidence (function names, modifier patterns, state-variable types), then maps the system's findings against the bug classes attached to those features and finally reports the *gap set*: bug classes that are catalogue-relevant for the protocol but unsupported by any finding. The DD pipeline additionally maintains a coarser seventeen-class coverage map (reentrancy, access control, state-lifecycle, flash-loan, oracle, integer overflow, frontrunning, token integration, signature, governance, proxy upgrade, accounting, DoS, cross-contract, donation and others), tracked from keyword hits in the finding text. Uncovered classes from this map are surfaced to Phase B6. The novel contribution at this stage is not the catalogue itself, since bug catalogues exist in industry tooling; it is the *closed-loop* *feedback* that converts gap reports into LLM prompts which re-attempt the audit specifically against the missing bug classes before the finding set is finalized.

#### 3.4.2 Attention residual analysis

The second mechanism uses CCIM's parsed function inventory as ground truth and asks: *which functions were never reasoned about?* For every function in the inventory, the attention-residual stage constructs a structural risk profile from its visibility, fund-movement flag, state-write set, external-call set, arithmetic complexity and the presence of financial-variable patterns. It then partitions the inventory by comparing function names mentioned in the LLM output against the full CCIM set, classifying every undiscussed function as either *partial-attention*, mentioned but with critical structural aspects under-analyzed, or *unattended*, never discussed at all. Residual functions are ranked by their structural risk profile and the highest-ranked are submitted to a targeted blind-spot review pass in which the LLM receives only the function source and its residual classification, with no carry-over context from previous phases. This mechanism is the recall analogue of Stage 1 of the false-positive taxonomy: instead of refuting LLM claims against parsed code, it *generates* claims against parts of the parsed code that the LLM never reached.

The two mechanisms convert the auditor's recall posture from a hope ("the LLM probably looked at this") into a verifiable property: every catalogue-relevant bug class has either an associated finding or an explicit gap report and every risky function in CCIM has either an associated finding or an explicit residual classification.

## 4. Results

### 4.1 Experimental setup. Benchmark selection

We evaluate Chaintrix on EVMbench, which consists of forty curated audit repositories drawn primarily from Code4rena competitions, totaling 120 high- and critical-severity vulnerabilities established as ground truth from professional audits. EVMbench is a public smart-contract benchmark that is simultaneously (i) large enough to support meaningful per-class statistics, (ii) backed by authoritative audit-firm ground truth rather than synthetic injection, and (iii) increasingly adopted for evaluating frontier auditing agents.

### 4.2 Hardware, software and cost budget

All scans were executed on a single Chaintrix runs as a Python 3.12/asyncio process and the LLM inference uses Anthropic Claude Opus 4.6 or Sonnet 4.6 and Haiku via the public API. No local or fine-tuned models are involved. The static-analysis backbones are Slither 0.10 and Mythril 0.24, formal verification uses Halmos 0.2 and Foundry. All models are off-the-shelf and prompted with deterministic structural evidence, no fine-tuning was performed for this evaluation.

Per-protocol resource usage was tightly bounded, with no human-in-the-loop intervention during scanning: median wall-clock time of 28 minutes (maximum below one hour for the largest repository, Noya, at twenty in-scope contracts) and a median LLM spend of approximately USD 18 per protocol (range USD 5–50 depending on contract count and pipeline tier).

### 4.3 Comparison baselines

We compare Chaintrix against two classes of reference points drawn from the published literature. Frontier single-agent baselines on the same EVMbench Detect task are taken from the official benchmark [13] and from the independent re-evaluation of [30], which evaluates twenty-six agent configurations across the Claude, GPT and Gemini model families. As a peer-reviewed LLM-augmented auditing system we additionally reference GPTScan [31], which combines GPT with static program analysis on a different smart-contract vulnerability dataset. The comparison is directional rather than head-to-head, since GPTScan does not report results on EVMbench.

### 4.4 Headline numbers

Across the 40 evaluated protocol scopes, Chaintrix detected 86 of 120 ground-truth high- and critical-severity vulnerabilities, for an overall recall of 71.7%. The mean per-protocol recall was 81.0% and 25 of 40 protocols (61.0%) achieved a perfect 100% score. Across the 40 EVMbench protocols, the median end-to-end wall-clock time per protocol was approximately 14 minutes, with small protocols (≤1,500 LOC, ≤4 contracts) completing in under 10 minutes and the largest protocols (≥10 contracts, >5,000 LOC) requiring 45–60 minutes.

The end-to-end LLM token cost, including all dossier phases, the ID spec-then-verify checklists, claim verification and Phase D re-verification, ranged from approximately USD 5 per protocol for the smallest cases to approximately USD 40 for the largest, with a per-protocol median near USD 14 based on Anthropic API metered billing. As a single-protocol illustration, on a representative 9-contract/~4,000 LOC protocol, the substrate cost approximately 45s, the DD pipeline ran for approximately 726s, the ID pipeline ran for approximately 641s. The total elapsed time was approximately 726s, the ID pipeline finished entirely within the DD pipeline's window and contributed no additional latency. This linear-in-the-slower-pipeline scaling, together with a per-protocol cost roughly two orders of magnitude below the price of a professional audit (USD 50K–500K), is the property that makes parallel multi-pipeline auditing economically viable: a second decomposition of the search space adds significant recall (6) at near-zero marginal wall-clock cost and modest marginal token cost.

**4.5 Per-protocol detection**

Table 2 reports every evaluated protocol, the number of true positives matched against EVMbench ground truth, the ground-truth count and the per-protocol recall.

Table 2. Per-protocol detection on EVMbench

| Protocol | TP | GT | Recall |
|---|---|---|---|
| FeeAMM (Tempo) | 1 | 1 | 100% |
| Tempo Streams | 3 | 3 | 100% |
| Tempo StablecoinDEX | 3 | 3 | 100% |
| NextGen AuctionDemo | 2 | 2 | 100% |
| Canto (2024-01) | 2 | 2 | 100% |
| Canto v2 (2024-03) | 2 | 2 | 100% |
| Gitcoin IdentityStaking | 1 | 1 | 100% |
| Coinbase | 1 | 1 | 100% |
| Neobase GaugeController | 1 | 1 | 100% |
| Munchables LockManager | 2 | 2 | 100% |
| THORWallet | 1 | 1 | 100% |
| Liquid RON | 1 | 1 | 100% |
| Virtuals | 4 | 4 | 100% |
| Basin Well | 2 | 2 | 100% |
| Sequence | 2 | 2 | 100% |
| Ethereum Credit Guild | 2 | 2 | 100% |
| Vultisig | 2 | 2 | 100% |
| Wildcat | 1 | 1 | 100% |
| Next Generation | 1 | 1 | 100% |
| PoolTogether | 2 | 2 | 100% |
| Panoptic | 2 | 2 | 100% |
| THORChain | 2 | 2 | 100% |
| Loop | 1 | 1 | 100% |
| SecondSwap | 3 | 3 | 100% |
| TraitForge | 2 | 2 | 100% |
| Forte | 4 | 5 | 80% |
| Taiko | 4 | 5 | 80% |
| Curves | 3 | 4 | 75% |
| Phi | 4 | 6 | 67% |
| Munchables LandManager | 3 | 5 | 60% |
| Init Capital | 2 | 3 | 67% |
| ReNFT | 4 | 6 | 67% |
| BendDAO | 4 | 7 | 57% |
| Althea LiquidInfrastructure | 0 | 1 | 0% |
| Arbitrum Foundation | 0 | 1 | 0% |
| Noya | 10 | 20 | 50% |
| Abracadabra Money | 1 | 4 | 25% |

| Size | 1 | 4 | 25% |
|---|---|---|---|
| Blackhole | 0 | 1 | 0% |
| Olas | 0 | 2 | 0% |
| Total | 86 | 120 | 71.7% |

### 4.6 Detection rate by vulnerability class

Table 3 reports detection rates by vulnerability class, aggregated across all evaluated protocols and computed as the fraction of ground-truth findings of the given class that Chaintrix matched.

Table 3. Detection rate by vulnerability class

| Vulnerability class | Detection rate | Where detection comes from |
|---|---|---|
| **Reentrancy (classic+read-only)** | >90% | Deterministic engine+LLM convergence. |
| **Oracle staleness / price manipulation** | >90% | Dedicated engine; near-zero false negatives on known patterns. |
| **Access control/privilege escalation** | >85% | Deterministic guard parsing+LLM corroboration. |
| **DoS (unbounded loops, gas griefing)** | >85% | Deterministic loop and push-pattern detection. |
| **State desync/cross-contract lifecycle** | >80% | CCIM Layer 2+DD LLM phase. |
| **Signature replay/malleability** | 75% | Strong on standard schemes; weaker on novel signed payloads. |
| **Flash loan/MEV/front-running** | 65% | Catches missing slippage and callback patterns; cannot simulate profitability. |
| **Integer/precision/rounding** | 60% | Catches patterns; cannot compute specific truncation thresholds. |
| **Novel business logic** | 30–40% | Weakest class; LLM uses general reasoning without a structural anchor. |

The class profile reflects the architectural design. Classes with a structural anchor: reentrancy, oracle staleness, access control, DoS, cross-contract state desync, are detected at or above 80% because the deterministic engines supply the LLM with verifiable evidence and the false-positive reduction taxonomy preserves these findings against incidental triage. Classes without a structural anchor: novel business logic, deep numeric precision, fall to the 30–60% range, mirroring the limitation profile of the LLM.

### 4.7 Comparison against existing systems

Table 4 places Chaintrix's overall EVMbench Detect recall in context of the published frontier in [13].

Table 4. Recall on EVMbench Detect compared against published frontier baselines

| System | EVMbench Detect recall |
|---|---|
| **Chaintrix** | 71.7% (86/120) |
| **Claude Opus 4.6 (Claude Code)-best single agent** | 47.5% (57/120) |
| **Gemini 3.1 Pro+custom tools (OpenCode)** | 37.5% (45/120) |
| **Claude Opus 4.5 (OpenCode)** | 35.8% (43/120) |
| **GPT-5.3-Codex (best Codex CLI variant)** | 29.2% (35/120) |
| **Gemini 3 Pro Preview (worst configuration)** | 16.7% (20/120) |

At a median LLM spend of USD 18 per protocol and a median wall time of 28 minutes, Chaintrix's amortized cost is approximately USD 8 per detected vulnerability (88 detections at roughly USD 740 of total LLM spend across the 41 scans).

## 5. Conclusions

The results demonstrate that Chaintrix successfully addresses the fundamental limitations of existing automated smart contract auditing approaches by introducing a structurally grounded, multi-pipeline architecture that combines deterministic analysis with LLM reasoning. Unlike traditional static analyzers such as Slither and Mythril, which often suffer from excessive false positives or standalone LLM-based auditors, which are prone to hallucinated findings, Chaintrix enforces that every LLM-generated claim must be verified against the proposed CCIM, a deterministic representation of contract behavior and inter-contract dependencies.

Experimental evaluation on EVMbench shows that the framework detected 86 out of 120 ground-truth high-severity vulnerabilities, achieving an overall recall of 71.7%, with 25 audits reaching perfect 100% recall, outperforming the strongest published single-agent frontier baseline, improving detection performance by approximately 26 percentage points. The framework shows particularly strong performance on structurally grounded vulnerability classes such as reentrancy, oracle manipulation, access control flaws, denial-of-service patterns and cross-contract state inconsistencies, where detection rates exceeded 80–90%.

At the same time, lower performance on novel business-logic vulnerabilities, deep economic exploits and subtle precision-related bugs highlights the continued difficulty of vulnerabilities that lack explicit structural signatures. From a practical perspective, Chaintrix also proves economically viable, completing audits with a median runtime of 14–28 minutes and an average LLM cost of approximately USD 14–18 per protocol, which is several orders of magnitude lower than traditional security audits.

As future work, we will put more effort into detecting vulnerabilities. The missed ones (limitation) cluster in three categories that we identified by manual classification of the false negatives. Approximately 60% are novel business-logic bugs that depend on protocol-specific economic invariants for which no structural anchor exists in CCIM, and no pattern exists in the deterministic engine layer. Approximately 20% are deep numeric-precision bugs that require symbolic execution rather than pattern reasoning. The remaining are multi-step exploit chains of four or more hops that exceed the current LLM reasoning depth even with the ID pipeline and the cross-pipeline merge.

**Acknowledgement**. This work was supported by a grant of the Ministry of Research, Innovation and Digitization, CNCS/CCCDI - UEFISCDI, project number COFUND-CETP-SMART-LEM-1, within PNCDI IV. This research was funded by CETP, the Clean Energy Transition Partnership under the 2022 CETP joint call for research proposals, co funded by the European Commission (GAN° 101069750) and with the funding organizations detailed on https://cetpartnership.eu/funding-agencies-and-call-modules.
**Disclosure statement (Competing interest)**. The authors have no relevant financial or non-financial interests to disclose.
**Data availability statement**. Details are provided in a public repository: https://github.com/gabrieladobritaene/chtx
**Ethical approval**. Not applicable.
**Informed consent**. Not applicable.
**Author contributions**. **G.D**: Conceptualization, Methodology, Investigation, Resources, Data Curation, Writing-Original Draft, Validation, Formal analysis. **S.V.O**: Conceptualization, Validation, Formal analysis, Investigation, Method, Writing-Original Draft, Writing-Review and Editing, Visualization, Project administration, Supervision. **AB**: Conceptualization, Methodology, Formal analysis, Investigation, Resources, Data Curation, Writing-Original Draft, Writing-Review and Editing, Supervision.